\documentclass[10pt,twocolumn,letterpaper]{article}

\usepackage{iccv}              

\usepackage{multirow}
\usepackage{makecell}
\usepackage{tablefootnote}

\definecolor{iccvblue}{rgb}{0.21,0.49,0.74}
\usepackage[pagebackref,breaklinks,colorlinks,allcolors=iccvblue]{hyperref}
\usepackage[export]{adjustbox}
\usepackage{bm} 
\usepackage{amsmath}
\def\paperID{6560} 
\def\confName{ICCV}
\def\confYear{2025}

\title{A Differentiable Wave Optics Model for End-to-End Computational Imaging System
Optimization}

\author{Chi-Jui Ho \quad Yash Belhe \quad Steve Rotenberg \quad Ravi Ramamoorthi \quad Tzu-Mao Li \quad Nicholas Antipa \\
University of California, San Diego\\
{\tt\small \{chh009, ybelhe, srotenberg, ravir, tzli, nantipa\}@ucsd.edu}}

\begin{document}
\maketitle
\begin{abstract}
End-to-end optimization, which integrates differentiable optics simulators with computational algorithms, enables the joint design of hardware and software for data-driven imaging systems. However, existing methods usually compromise physical accuracy by neglecting wave optics or off-axis effects due to the high computational cost of modeling both aberration and diffraction. This limitation raises concerns about the robustness of optimized designs. In this paper, we propose a differentiable optics simulator that accurately and efficiently models aberration and diffraction in compound optics and allows us to analyze the role and impact of diffraction in end-to-end optimization. Experimental results demonstrate that compared with ray-optics-based optimization, diffraction-aware optimization improves system robustness to diffraction blur. Through accurate wave optics modeling, we also apply the simulator to optimize the Fizeau interferometer and freeform optics elements. These findings underscore the importance of accurate wave optics modeling in robust end-to-end optimization. Our code is publicly available at: \url{https://github.com/JerryHoTaiwan/DeepWaveOptics}
\end{abstract}    
\vspace{-6mm}
\section{Introduction}
\label{sec:intro}

The interdependence between optics and downstream algorithms is pivotal in imaging system design. 
To leverage this interdependence and achieve joint designs, end-to-end differentiable models, which incorporate a differentiable optics simulator and a computer vision algorithm, have been applied to simultaneously optimize hardware and software across a range of vision tasks \cite{cote2023differentiable, yang2023image, tan2021codedstereo, yang2023aberration, sitzmann2018end, peng2019learned, sun2021end, shi2022seeing}.
Given an image, the differentiable simulator models the corresponding measurement taken by the optics system, and the computer vision algorithm extracts semantic information. With a differentiable simulator and algorithm, a loss function scores task performance and drives the optimization of the optics and algorithm parameters via backpropagation.

A notable challenge in end-to-end optimization is incorporating wave optics effects in large field-of-view (FoV) and analyzing how the fidelity of optics simulation impacts overall system optimization. Realistic modeling requires accounting for diffraction across the entire sensor, which is computationally expensive. Thus, many designs neglect diffraction and adopt ray optics \cite{cote2023differentiable, sun2021end, yang2023image}. Some simulators consider simplified diffraction using thin-phase surfaces \cite{shi2022seeing, wei2023modeling} or shift-invariance \cite{sitzmann2018end, baek2021single, he2023learned}, limiting applicability to multi-element or compound optics. Although recent frameworks model more realistic wave optics \cite{chen2021optical, yang2024end}, their accuracy and efficiency in different configurations remain questionable, and the significance of wave optic effects on system optimization remains an open problem. 

\begin{figure}[t]
     \centering
     \includegraphics[width=0.48\textwidth]{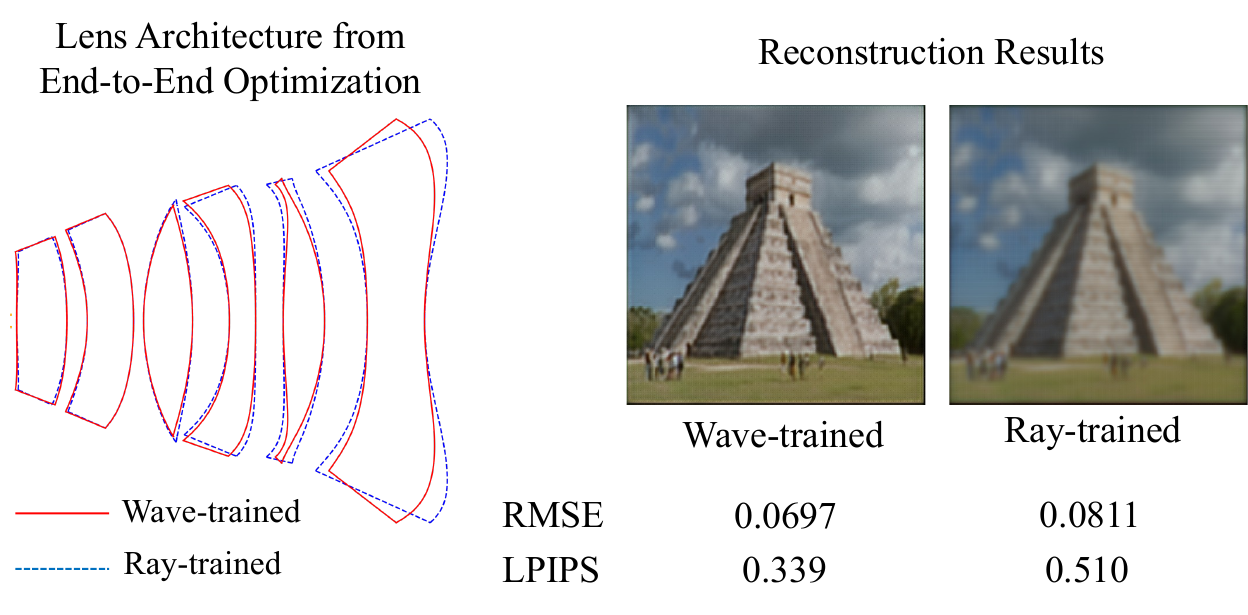}
     \vspace{-5mm}
     \caption{\textbf{End-to-end optimized lens architectures and reconstruction models using ray and wave optics.} By taking diffraction into account, our wave-trained model yields sharper reconstruction results than the baseline using ray optics.}
     \vspace{-6mm}
     \label{fig:teaser}
\end{figure}

In this paper, we propose an accurate, efficient, and differentiable optics simulator, which uses ray tracing with the Rayleigh-Sommerfield integral \cite{goodman2005introduction} to model diffraction and off-axis aberrations in compound optical systems without thin-phase or paraxial approximations. To effectively model diffraction in large FoVs, we use an interpolation method to approximate the measurements with a subset of point spread functions (PSFs). By providing accurate and efficient wave optics rendering, the proposed simulator enables us to incorporate diffraction into end-to-end optimization and analyze its role and impact on imaging system design.

Unlike systems optimized solely under ray optics assumptions, our wave optics model guides the system to a design with reduced diffraction effects. An example of lens architecture and system performance optimized by ray and wave optics is shown in Fig. \ref{fig:teaser}.

Our contributions are 
\begin{itemize}
    \item We propose a differentiable model that accurately accounts for aberration and diffraction in compound optical systems. With efficient rendering, the model is compatible with end-to-end optimization.
    \item We analyze the role of diffraction in end-to-end design. Neglecting diffraction leads to suboptimal lens and algorithm configurations. Conversely, by accurately modeling diffraction, our model attains superior solutions. 
    \item The proposed simulator is applicable to a wide range of wave-optics-based imaging systems, including interferometric setups and freeform optical systems.
\end{itemize}

\section{Related Work}
\label{sec:related_work}

\subsection{End-to-End Optimization}

Conventional lens design optimizes a merit function that combines lens properties with transfer function quality \cite{zemax}. However, such merit functions do not necessarily correlate with computer vision task performance \cite{fontbonne2022comparison, yang2023image}.  End-to-end optimization addresses this by directly optimizing task performance, jointly refining hardware and software. Leveraging differentiable optics simulators and inference algorithms on large datasets \cite{sun2021end, cote2023differentiable, yang2023image, he2023learned}, this approach provides data-driven designs that capture interdependencies among optics, algorithms, and tasks \cite{fontbonne2022comparison}. 

This paradigm has been applied to image reconstruction \cite{sitzmann2018end, peng2019learned, shi2022seeing, fontbonne2022comparison, liu2021end, li2021end} and restoration \cite{hale2021end, zhang2023end}. Sitzmann \textit{et al.} extend the depth of field in computational cameras \cite{sitzmann2018end}, while Peng \textit{et al.} achieve high-FoV image reconstruction \cite{peng2019learned}. Shi \textit{et al.} combine diffractive optics with point-PSF-aware neural networks to recover occluded scenes \cite{shi2022seeing}. Beyond reconstruction, this strategy has also been applied to semantic tasks: Baek \textit{et al.} jointly optimize diffractive elements and networks for hyperspectral depth sensing \cite{baek2021single}; Kellman \textit{et al.} design coded-illumination patterns and unrolled networks for phase recovery \cite{kellman2019physics}; Pidhorskyi \textit{et al.} develop a differentiable ray tracer for depth-of-field-aware intensity recovery \cite{pidhorskyi2022depth}; Yang \textit{et al.} optimize off-axis aberrations for image classification \cite{yang2023image}; and Cote \textit{et al.} co-optimize lens materials and structures for object detection \cite{cote2023differentiable}. These works demonstrate how end-to-end optimization yields task-specific optics–algorithm co-design.

\subsection{Balancing Accuracy and Efficiency in Differentiable Optics Simulation }\label{subsec:physics}

Zemax is an industry-standard tool for accurate wave‑optics modeling using Huygens’ principle, but its slow computational speed restricts its use in end‑to‑end optimization \cite{zemax}, which requires gradient propagation to model complex optical–semantic relationships  \cite{wang2022differentiable}.
To mitigate the cost in large FoV differentiable rendering, simplified physics models are usually adopted, such as thin-lens modeling \cite{peng2019learned, shi2022seeing} and geometric ray tracing \cite{sun2021end, cote2023differentiable, wang2022differentiable}, to compute PSFs in end-to-end optimization. However, the former is limited to a single thin lens, and the latter neglects wave effects.  

It is also common to model simplified wave-optical effects. Sitzmann \textit{et al.} use Fresnel propagation for shift-invariant diffractive optics \cite{sitzmann2018end}; He \textit{et al.} compute PSFs with diffraction theory in shift-variant systems \cite{he2023learned}; and Tseng \textit{et al.} replace the full pipeline with a neural PSF renderer \cite{tseng2021differentiable}. All these assumptions limit their applicability to compound optical systems. Chen \textit{et al.} \cite{chen2021optical} simulate diffraction with ray tracing but neglect magnitude variations with propagation distance, which become inaccurate when modeling defocus. Wei \textit{et al.} \cite{wei2023modeling} and Yang \textit{et al.} \cite{yang2024end} use the angular spectrum method (ASM) for free-space propagation, which requires high sampling density when modeling defocus. A similar limitation arises in field tracing \cite{wyrowski2011introduction}, which unifies different propagation models but still relies on ASM, while the generalized Debye integral \cite{wang2020generalized} accelerates focal-field computation via a homeomorphic Fourier transform yet remains constrained to low Fresnel number regimes. These considerations motivate our differentiable ray–wave framework, which explicitly models diffraction from rays and examines the impact of accurate wave modeling on end-to-end optimization.
\begin{figure*}[t]
    \centering
    \includegraphics[width=\textwidth]{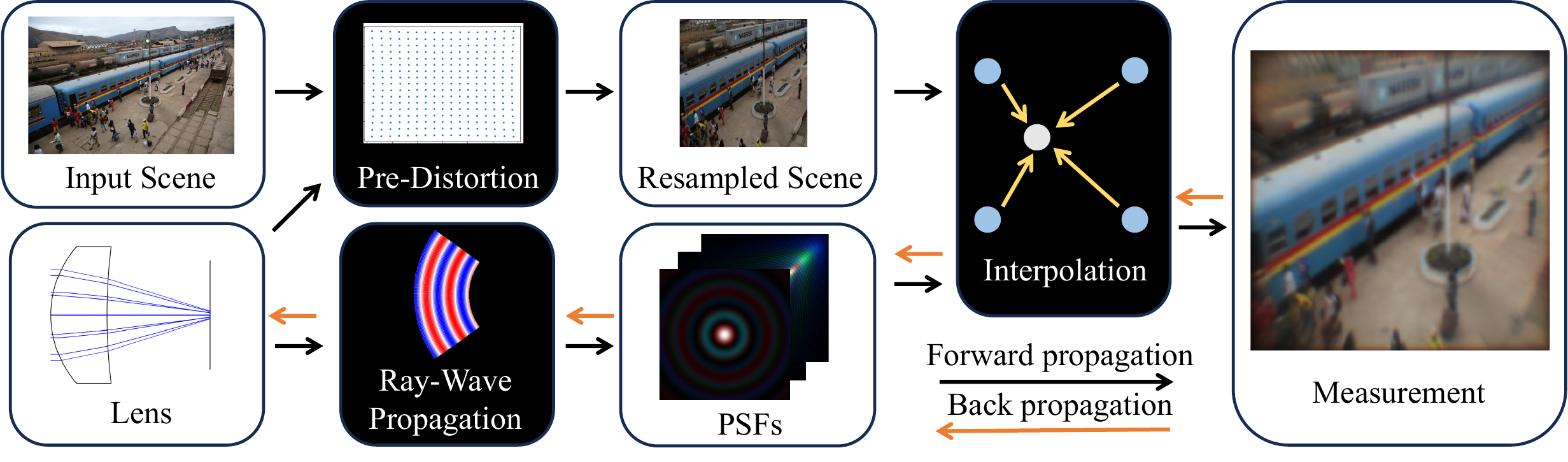}
    \vspace{-4mm}
    \caption{\textbf{Our proposed differentiable wave optics simulator.} Given an input scene and lens configuration, we first resample the scene based on the lens' pre-distortion map. Next, we generate diffraction-aware PSFs using our wave optics simulator. Finally, we interpolate the convolution of the resampled scene with the PSFs to obtain our final measurement. During lens optimization, measurement gradients are back-propagated to the lens parameters.}
    \vspace{-5mm}
    \label{fig:overview}
\end{figure*}

\section{Differentiable Optics Model}
\label{sec:diff_wave_model}

Our differentiable simulator is designed to accurately and efficiently capture both aberration and diffraction in compound optics systems, making it a robust rendering model and providing scalable end-to-end optimization. 

An overview of our differentiable hybrid ray-wave imaging simulator is shown in Fig. \ref{fig:overview}. Given a point light source at $\mathbf{x} = (x, y, z)$ and an optical system with sequential refractive surfaces, our model incorporates a differentiable ray tracer \cite{wang2022differentiable} and Rayleigh-Sommerfield integral \cite{goodman2005introduction} to account for wave optics effects in PSF $h(\mathbf u|\mathbf x)$, where $\mathbf u$ denotes sensor pixel position. We describe PSF rendering in detail in Sec. \ref{subsec:waveoptics}. Furthermore, given scene intensity $b(\mathbf{x})$, the resulting measurement $I(\mathbf{u})$ is derived from the superposition integral of incoherent PSFs \cite{chen2021optical}:

\begin{equation}
    I(\mathbf u) = \int b(\mathbf x) h(\mathbf u | \mathbf x) d\mathbf x.
    \label{eq:general}
\end{equation} 
 
Directly computing Eq.  \eqref{eq:general} across the full FoV is computationally intensive, requiring full-resolution PSF rendering for every point source. To address this, we develop an efficient interpolation method that balances accuracy and computational cost. The approach involves sampling a subset of PSFs and using interpolation to approximate the full measurement by convolving the subset PSFs with their corresponding sub-scene intensities \cite{baktash2022computational}  (details in Sec. 
\ref{subsec:approx}). By efficiently modeling diffraction, off-axis aberrations, and geometric distortions, this approach enhances the robustness of data-driven lens design and enables deeper analysis of wave optics effects in imaging systems.

\subsection{PSF Rendering}
\label{subsec:waveoptics}

\begin{figure}[t]
    \centering
    \includegraphics[width=0.5\textwidth]{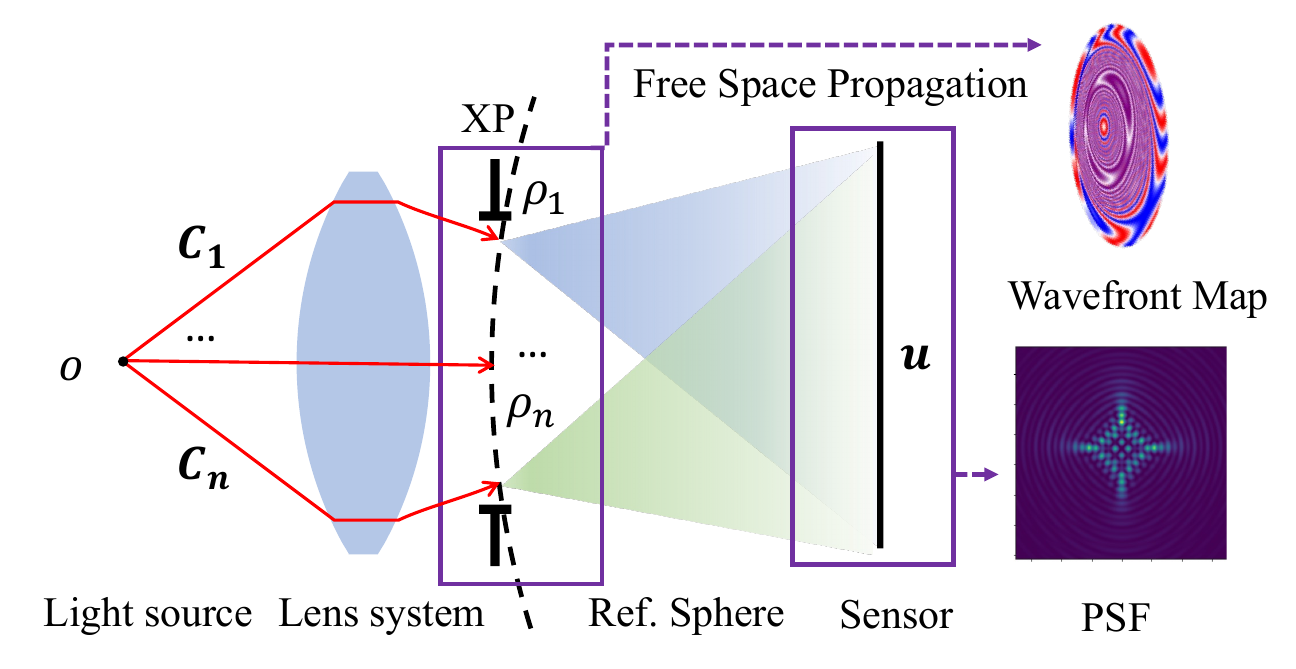}
    \vspace{-6mm}
    \caption{\textbf{Our wave optics simulator.} We trace rays emitted from a point source $o$ to the reference sphere on the system's exit pupil, and compute intersections $\{\bm \rho_i\}$ and associated phase on a wavefront map. We then perform free-space propagation toward the sensor to generate a PSF. XP: Exit Pupil. Ref: Reference. }
    \vspace{-3mm}
    \label{fig:ray tracing }
\end{figure}

A conceptual flow of our wave optics model is illustrated in Fig. \ref{fig:ray tracing }. 
To compute a PSF, we first use geometric ray tracing to sample the wavefront map in the exit pupil, and then propagate the complex field of the wavefront map to the sensor plane. 
In ray tracing, we use Newton's Method \cite{sun2021end, wang2022differentiable} to calculate the intersections between rays and surfaces and use Snell's Law to model refractions. The wavefront map is then calculated in the reference sphere, whose center and radius are determined by the intersection between the principal ray (the ray that passes through the pupil center) and the sensor plane, and the distance between the exit pupil (XP) and the sensor, respectively. Specifically, we approximate the XP using paraxial rays across all incident angles, consistent with prior work \cite{li2021end, yang2024end, zemax}.

 \begin{figure}[t!]
    \centering
    \includegraphics[width=0.20\textwidth]{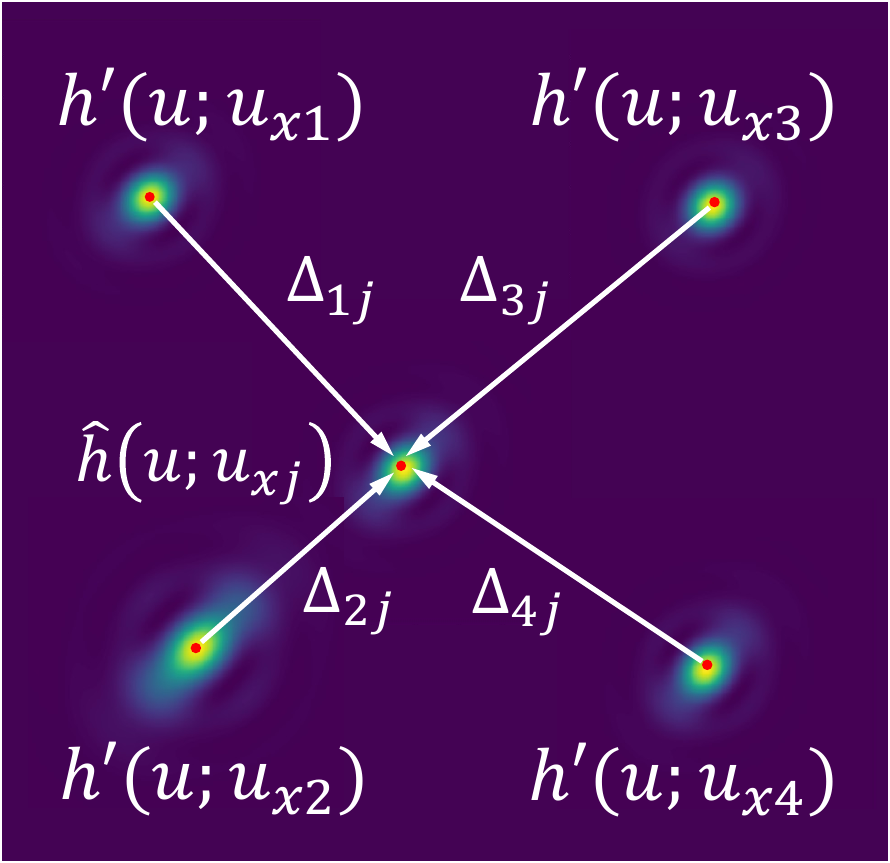}
    \caption{\textbf{Approximating unsampled PSFs.} Our system first samples PSFs $h'(u; \cdot)$ on a regular grid. Next, by exploiting the isoplanatic property, it approximates off-grid PSFs $\hat{h}(u; \cdot)$ by interpolating shifted and scaled versions of nearest samples $h'(u; \cdot)$. }

    \vspace{-6mm}
    \label{fig:isoplanatic}
\end{figure}

The task thus reduces to computing the amplitude and phase of the complex field on the reference sphere. Since the exit pupil is an image of the aperture stop, we model the amplitude by the square root of the aperture stop transmittance. The phase at the reference sphere is determined by the optical path length (OPL) $\delta$ calculated by
\begin{equation}
    \delta = \int_C n(s)ds {,}
    \label{eq:opl}
\end{equation}
where $n(s)$ is the 3D refractive index of the system and $C$ is the path that a given ray takes from the light source to the reference sphere \cite{jenkins1957fundamentals}. 

It is notable that the phase of complex values across the reference sphere, called the \textit{wavefront error map}, reflects the degree of focusing \cite{shannon1997art}. When the system is in-focus, the reference sphere exactly matches the wavefront, and the phase is constant on the sphere. Otherwise, the mismatch between the actual wavefront and the reference sphere causes phase variations across the reference sphere. Moreover, compared with the planar pupil field used by ASM-based modeling \cite{wei2023modeling, yang2023aberration}, the spherical structure effectively reduces the phase variation and hence alleviates the sampling requirement. In other words, we choose the reference sphere to model the wavefront error map because of its interpretability, efficiency, and compatibility with our propagation model, but the choice of the reference geometry is arbitrary and depends on the propagation model  
\cite{shannon1997art, zemax}.  

Consequently, for a ray piercing the reference sphere at $ \bm{\rho_i} = (\rho_{x_i}, \rho_{y_i}, \rho_{z_i})$, we model the complex field by 
\begin{equation}
    v(\bm \rho_i) = a_i \exp{(jk \delta_i)},
\label{eq:wavefront}
\end{equation}
\noindent where $a_i$ is the amplitude, $k$ is the wave number, $j=\sqrt{-1}$, and $\delta_i$ is the optical path length.

As shown in Fig. \ref{fig:ray tracing }, the propagation from the reference sphere to the sensor is in free space. The total intensity, $h(\textbf{u})$, at sensor coordinate $\textbf{u}$ is computed by the Rayleigh-Sommerfeld integral \cite{goodman2005introduction}, which we Monte-Carlo evaluate with $N$ coherent rays by

\begin{equation}
\label{eq:huygens}
    h(\textbf{u}) = \frac{1}{N\lambda^2} \Bigg|\sum_{i=1}^N v({\bm \rho_i})\frac{\exp{(jk|\vec{r}_{u,i} | )}}{|\vec{r}_{u,i} |}\cos(\theta_{u, i})\Bigg|^2,
\end{equation}

\noindent where $\vec{r}_{u,i} $ denotes the vector from $\bm \rho_i$ to sensor coordinate $\textbf{u}$, and $\theta_{u, i}$ is the angle between $\vec{r}_{u,i} $ and the normal vector of the reference geometry at $\bm \rho_i$. To accelerate computation, we vectorize the operations; however, because full vectorization can exceed memory limits, we apply checkpointing in PyTorch \cite{paszke2017automatic} to alleviate this issue.

 \begin{figure*}[ht!]
    \centering
    \includegraphics[width=0.92\textwidth]{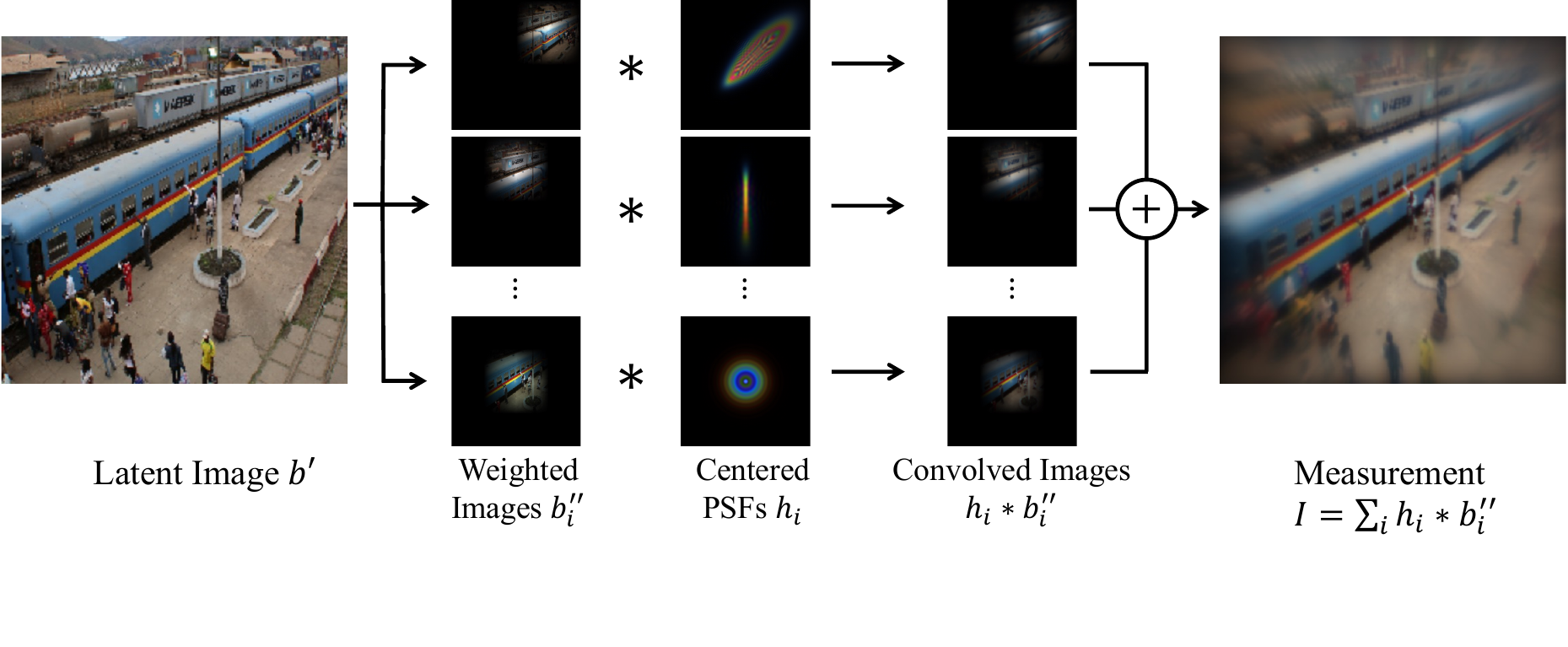}
    \vspace{-13mm}
    \caption{ \textbf{Rendering measurement with a subset of PSFs.} Given a latent image $b'$, we first generate weighted images $b_i''$. Next, we generate PSFs $h_i$ at the centers of weighted images and pair them with corresponding PSFs. Finally, we convolve weighted images and PSFs ($h_i*bi''$) and sum them up to obtain the measurement $I$.}
    \vspace{-6mm}
    \label{fig:fftconv}
\end{figure*}

\subsection{Approximating Superposition Integral}
\label{subsec:approx}

Although we can render PSFs with wave optics effects, the high computational costs make it challenging to exhaustively compute all PSFs. 
A common way to alleviate this cost is to assume the system is shift-invariant and approximate Eq. \eqref{eq:general} with a single convolution between the on-axis PSF and scene intensities \cite{goodman2005introduction}. However, this assumption is overly restrictive as it does not model common off-axis aberrations such as coma, astigmatism, and field curvatures. 

Therefore, we assume that PSFs are locally \textit{isoplanatic}; the system is shift-invariant over a sufficiently small area. This allows us to sample a small subset of PSFs and approximate the superposition integral through a sequence of convolutions, thereby saving computational costs while maintaining the ability to model off-axis aberrations. 

To facilitate the derivation, we parameterize scene intensities $b(\mathbf{x})$ and PSFs $h(\mathbf{u;x})$ in terms of sensor coordinates $\{\mathbf{u}\}$ as follows. Given a world coordinate $\mathbf{x}$ and lens distortion function $d(\cdot)$, we compute the intersection $\mathbf{u_x}=d(\mathbf{x})$ between the non-paraxial principal ray emitting from $\mathbf{x}$ and the sensor plane. Because the function is one-to-one,  $b(\mathbf{x})$ and $h(\mathbf{u;x})$ can be re-parameterized as $b'(\mathbf{u_x})$ and $h'(\mathbf{u;u_x})$, respectively. An example of distorted coordinates is visualized in Fig. \ref{fig:overview}. Because the distortion function $d(\cdot)$ only determines the input scene content, we only consider it in the inference, but not back-propagation. 

Fig. \ref{fig:isoplanatic} shows an example of approximating a PSF originating from an \textit{unsampled} world coordinate $\mathbf{x_j}$ according to PSFs $\{h(\mathbf{u} ; \mathbf{u_{xi}})\}$ originating from sampled world coordinates $\{\mathbf x_i\}$. For an unsampled PSF centered at $\mathbf {u_{xj}}$, we model it as the weighted sum of the known neighboring PSFs, which are aligned to the same location:

\begin{equation}
    \widehat{h}(\mathbf{u}; \mathbf{u_{xj}})=\sum_i w_i(\mathbf{u_{xj}}) h'(\mathbf{u}-\Delta_{ij}; \mathbf{u_{xi}}),
    \label{eq:neighor_psf}
\end{equation}

\noindent where $\Delta_{ij}=\mathbf{u_{xj}}-\mathbf{u_{xi}}$ is the center-to-center distance, in the sensor space, between the sampled PSF $i$ and unsampled PSF $j$.  $w_i(\mathbf{u_{xj}})$ determines the weight of the sampled PSF $i$ when approximating the unsampled PSF centered at  $\mathbf{u_{xj}}$.

Therefore, we rewrite Eq. \eqref{eq:general} by substituting the general form for the shift-varying PSFs found in Eq. \eqref{eq:neighor_psf}:

\begin{align}
    I(\mathbf u) &= \sum_{\mathbf{u_x}} b'(\mathbf{u_x}) \sum_i w_i(\mathbf{u_x}) h'(\mathbf{u}+\mathbf{u_{xi}}-\mathbf{u_x}; \mathbf{u_{xi})} \nonumber \\
     &= \sum_i \sum_{\mathbf{u_x}}  b''_i(\mathbf {u_{x}}) h'(\mathbf{u}+\mathbf{u_{xi}}-\mathbf{u_x}; \mathbf{u_{xi}}) {,}
    \label{eq:shift_u}
\end{align}

\noindent where $b''_i(\mathbf {u_x})=b'(\mathbf{u_x})w_i(\mathbf{u_x})$ represents the weighted latent image, which consists of input scene intensities distorted by the lens distortion curve and weighted by $w_i(\cdot)$. 

We observe that Eq. \eqref{eq:shift_u} is a sum of convolutions between the shifted version of sampled PSFs and the corresponding weighted latent image:

\begin{align}
    I(\mathbf u) &= \sum_i \sum_{\mathbf{u_x}} b''_i(\mathbf {u_x}) h_i(\mathbf u - \mathbf{u_x}) \nonumber \\
    &= \sum_i b''_i *h_i
    \label{eq:sum_u}
\end{align}

\noindent where  $h_i(\mathbf{u})=h'(\mathbf{u+u_i; u_{xi}})$.
Fig. \ref{fig:fftconv} illustrates an example of how we pair weighted images and PSFs, convolve them with each other, and sum up the convolved images to compute the measurement. Notably, because $v({\bm \rho_i})$ is obtained by differentiable ray tracing \cite{wang2022differentiable}, and the operations from \eqref{eq:huygens} to \eqref{eq:sum_u} are all differentiable, the entire pipeline remains differentiable. This property enables precise modeling of how lens configurations interact with wave‑optics effects to produce the measurements.

Although Zemax also uses Huygens' Principle to model wave propagation and serves as an industrial level baseline \cite{zemax}, it requires on-grid sampling to model the wavefront map, which limits the efficiency. The differentiability of Zemax is also limited to the given merit functions, while our simulator can be integrated with arbitrary differentiable algorithms. In the subsequent section, we incorporate this differentiable wave optics simulator into computer vision algorithms, allowing analysis of the impact of wave optics effects on optical systems tailored for vision tasks.

\section{Experiments}
\label{sec:exp}

With the simulator, we conduct joint optimization of optics systems and scene reconstruction algorithms, with a focus on analyzing the role of diffraction in end-to-end optimization. To the best of our knowledge, it is an unexplored experimental flow to analyze the requirements of physical accuracy in end-to-end optimization. We also analyze the rendering and interpolation accuracy of our simulator and extend the simulator to interferometry and freeform optics.

\begin{figure}[!t]
     \includegraphics[width=0.49\textwidth, left]{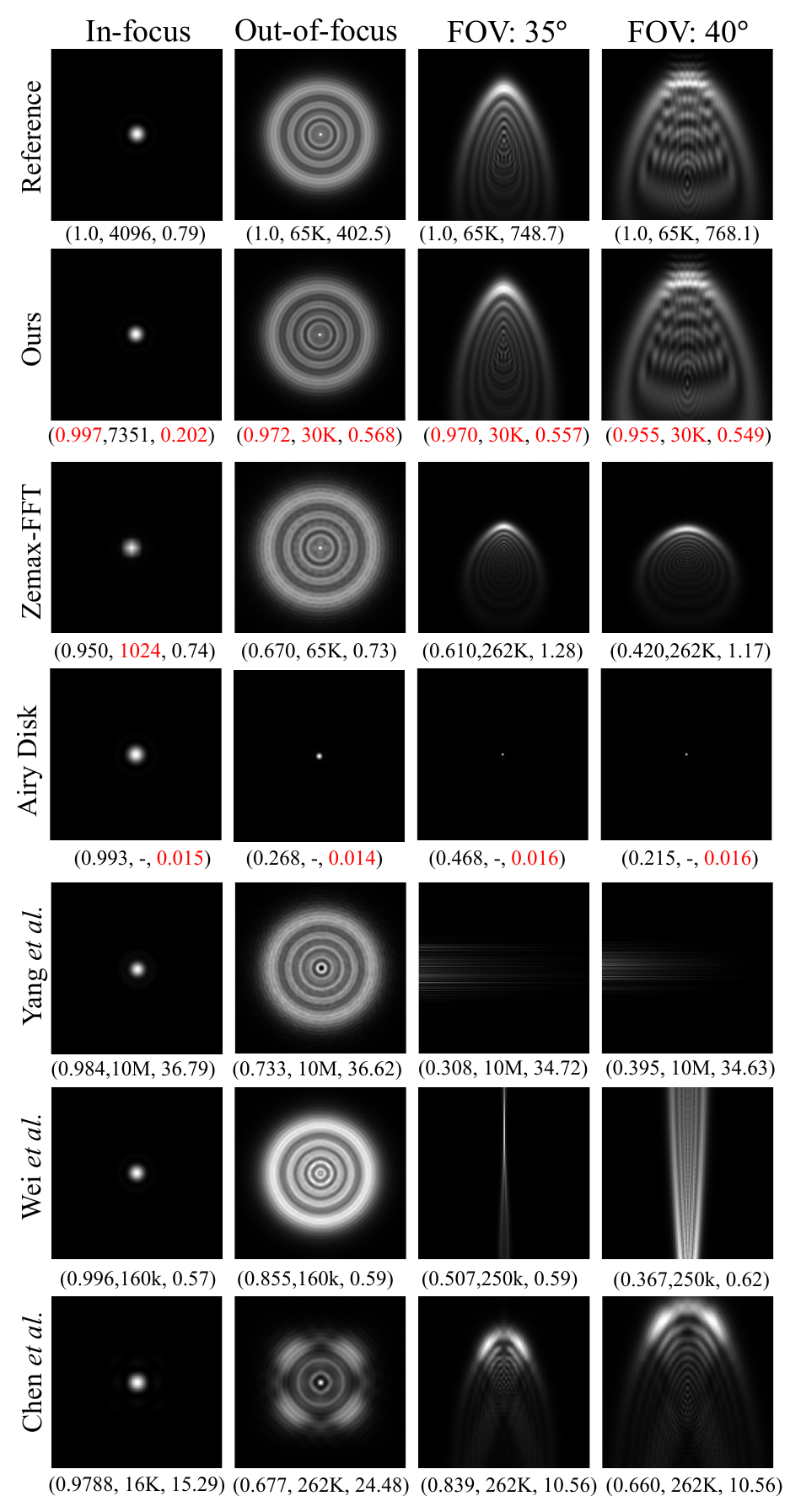}
     \vspace{-6mm}
     \caption{\textbf{PSFs rendered by different simulators under different conditions.} Unlike existing simulators \cite{yang2024end, chen2021optical, zemax, wei2023modeling}, ours avoids wavefront discretization and remains robust to defocus and large FoVs, achieving the highest accuracy and efficiency. The tuple (SSIM, ray count, time in sec.) highlights the best performance in red. As the Airy disk does not use ray-tracing, we skip its ray count and do not compare its time with others. Zoom in for details.}
     \vspace{-6mm}
     \label{fig:psf_cmp}
\end{figure}

\subsection{PSF Rendering}

In Fig. \ref{fig:psf_cmp}, we present monochromatic PSFs (wavelength: 532 nm) generated by our simulator alongside those from existing methods \cite{yang2024end, chen2021optical, wei2023modeling, zemax} under various conditions: On-axis PSFs for an in-focus and out-of-focus Cooke Triplet lens, and off-axis PSFs at 35° and 40° from a singlet lens. Because Zemax computes Huygens PSFs using the Rayleigh–Sommerfeld integral \cite{zemax, goodman2005introduction}, the most general scalar diffraction model, we use Zemax‑Huygens results as the reference. For each method, we report similarity to this reference using the structural similarity index (SSIM) and evaluate efficiency with ray count and computational time.

Overall, our method achieves superior accuracy and efficiency. While our simulator requires more rays than Zemax-FFT in the in-focus case, it attains higher accuracy with shorter runtime, underscoring both precision and efficiency. Notably, ASM-based methods \cite{yang2024end, wei2023modeling} are sensitive to defocus: as defocus increases, phase variations across the pupil plane and propagation kernel become extremely rapid. Since ASM discretizes these on a 2D grid, it struggles to capture such variations, reducing accuracy and efficiency. In contrast, our renderer supports off-grid wavefront maps that directly represent ray distributions, enabling efficient modeling of wave propagation. Although \citet{chen2021optical} also allows flexible ray distributions, their wave model does not account for magnitude changes from $|\vec{r}_{u,i}|$ in Eq.~\ref{eq:huygens}, instead projecting $|\vec{r}_{u,i}|$ onto ray directions. This approximation fails to capture magnitude variations across large spot sizes under defocus. Furthermore, defocus makes the Airy disk, commonly used to evaluate diffraction in ideal lenses, unreliable for modeling wave effects. Our results demonstrate that the proposed simulator is more robust, accurate, and efficient for defocused and large-FoV systems, conditions frequently encountered in end-to-end optimization. Importantly, our method is not merely reproducing the Huygens PSFs from Zemax but is more efficient and directly compatible with differentiable algorithms.

\subsection{System Optimization Setup}

In our imaging rendering process, we simulate beam propagation across the red, green, and blue light channels, compute the corresponding measurements for each wavelength, and then apply the Bayer filter to subsample these measurements. This results in blurred and mosaicked data.

We perform both ray-based and wave-based end-to-end optimization to jointly design lens systems and a U-Net \cite{ronneberger2015u} for scene reconstruction from system measurements. To compare their robustness to diffraction effects, we use wave optics in the evaluation. Input scenes are drawn from the DIV2K dataset \cite{agustsson2017ntire}, and lens configurations include variations in aperture radii and complexity, encompassing singlet, triplet, and six aspheric lenses. 

For optimization, we utilize the Adam optimizer \cite{kingma2014adam} to adjust both the network and lens parameters. The loss function includes root-mean-square error (RMSE) and perceptual loss (LPIPS) \cite{zhang2018unreasonable} between the normalized input scene intensities and the reconstructed results. To keep a consistent FoV for fair comparisons, whenever the focal length varies, we adjust the sensor size accordingly.

In addition to assessing reconstruction with RMSE and LPIPS, we use two metrics to quantify the disparity between ray- and wave-trained lenses: The mismatch between their F-numbers (MF) and the relative root mean squared error (RRMSE) of optimizable variables. All experiments were implemented on an Nvidia A40 GPU using PyTorch \cite{paszke2017automatic}. 

\begin{figure}[!t]
     \centering
     \includegraphics[width=0.48\textwidth]{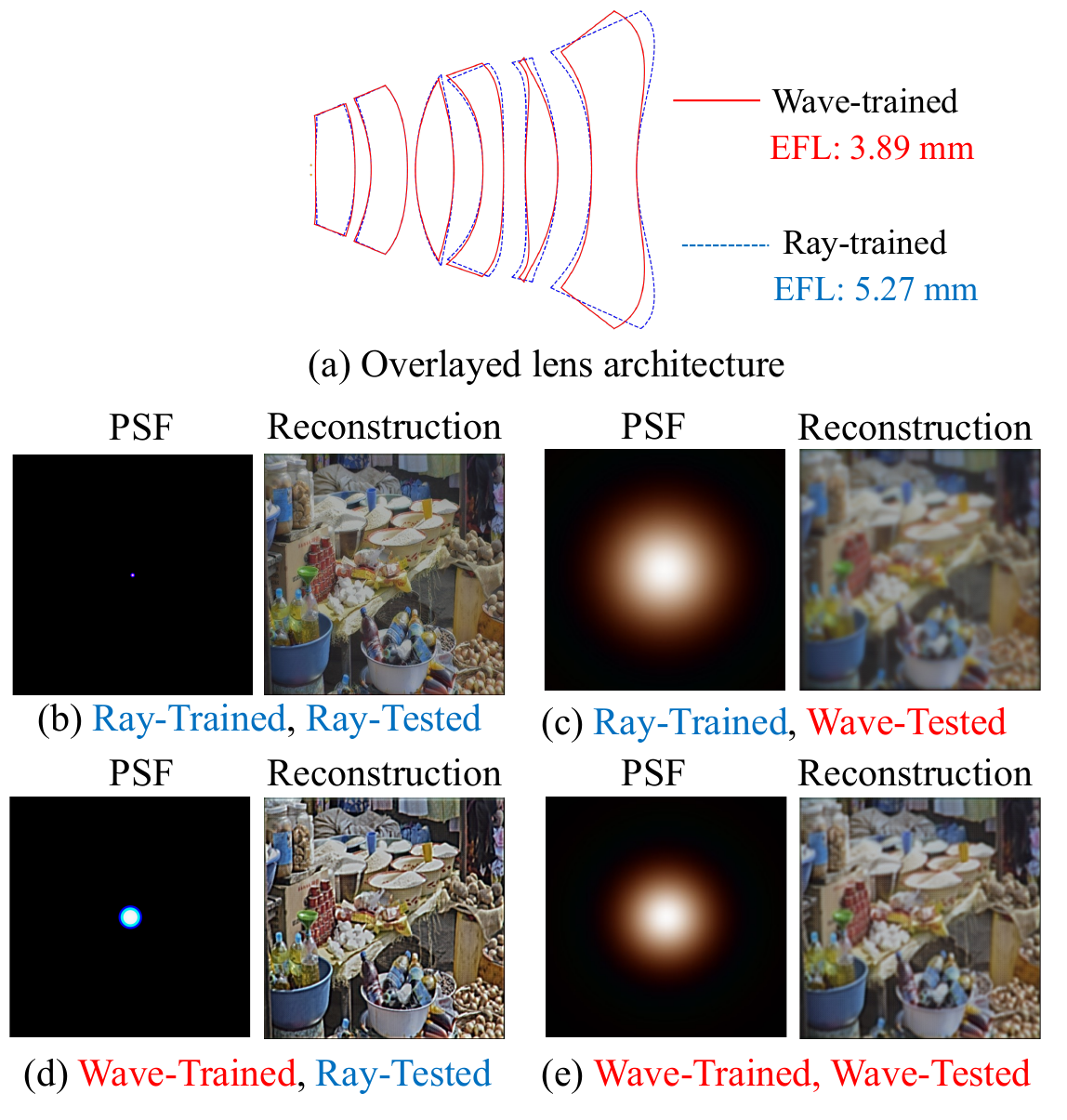}
     \vspace{-7mm}
     \caption{\textbf{Ray- vs. wave-trained systems.} (a) Lens architecture of wave- and ray-trained systems. The ray-trained system minimizes geometric spot size (b) but neglects diffraction blur (c). The wave-trained system has a larger geometric spot size (d), but a lower effective focal length (EFL) to control diffraction, yielding better reconstruction (e). PSF size: $0.044 \text{ mm}^2$.}
     \vspace{-6mm}
     \label{fig:results}
\end{figure}

\subsection{Demosaicking and Reconstruction}
\label{subsec:recon}

We summarize the reconstruction results in Table \ref{table:recon}, which are consistently evaluated using wave optics. Notably, with lenses having a 0.1 mm aperture radius, wave-training and ray-training yield different configurations and reconstruction performance. In Fig. \ref{fig:results}, we visualize ray- and wave-trained lens configurations and associated PSFs and reconstructions at different testing situations. As shown in Fig. \ref{fig:results} (a) and (b), the wave-trained lens changes its architecture to shorten the focal length and weaken diffraction, while the ray-trained lens focuses on minimizing RMS spot size.

Although the ray-trained system achieves a smaller geometric spot size, as shown in Fig. \ref{fig:results} (b) and (d), it fails to account for diffraction blur. When evaluated by accurate wave modeling, as shown in Fig. \ref{fig:results} (c) and (e), both PSF quality and reconstruction performance degrade. In contrast, while the wave-trained system slightly sacrifices geometric spot size, its optimized lens architecture effectively mitigates diffraction, the actual PSF-limiting factor, enhancing diffraction-limited resolution and producing sharper reconstructions. This highlights the critical role of diffraction in end-to-end optimization and the risks of neglecting it.

\begin{table}[t!]
\setlength{\tabcolsep}{4pt} 
\caption{Reconstruction performance on wave-optics-rendered measurements (RMSE / LPIPS) and lens disparity.}
\vspace{-2mm}
\centering
\begin{tabular}{c|cc|c|c}
\hline
\multirow{2}{*}{AR} & \multicolumn{2}{c|}{Training physics}    & \multirow{2}{*}{MF} & \multirow{2}{*}{RRMSE} \\ 
\cline{2-3} & \multicolumn{1}{c|}{Wave} & Ray &  \\
\hline
\multicolumn{5}{c}{Singlet Lens}\\
\hline
0.1 & \multicolumn{1}{c|}{\textbf{0.075 / 0.181}} & 0.089 / 0.451  & 1.11 & 5.1$\times 10^{-3}$ \\
0.3 & \multicolumn{1}{c|}{\textbf{0.065} / 0.076} & 0.063 / \textbf{0.073} & 0.108 & 6.8$\times 10^{-4}$\\ 
\hline
\multicolumn{5}{c}{Cooke Triplet Lens}\\
\hline
0.1 & \multicolumn{1}{c|}{\textbf{0.106 / 0.265}} & 0.148 / 0.772 & 8.689 & 0.580\\
0.3 & \multicolumn{1}{c|}{\textbf{0.104 / 0.230}} & 0.112 / 0.483 & 0.073 & 4.8$\times 10^{-3}$\\ 
\hline
\multicolumn{5}{c}{Six Aspherical Lenses}\\
\hline
0.1 & \multicolumn{1}{c|}{\textbf{0.085 / 0.368}} & 0.104 / 0.604 & 6.873 & 0.263 \\
0.3 & \multicolumn{1}{c|}{\textbf{0.067 / 0.173}} & 0.071 / 0.242 & 0.432 & 0.060\\ 
\hline
\multicolumn{5}{l}{\footnotesize AR: Aperture radius (unit: mm)} \\
\end{tabular}
\vspace{-4mm}
\label{table:recon}
\end{table}


Table \ref{table:recon} also shows that increasing the aperture radius from 0.1 to 0.3 mm reduces the mismatch between lens designs and the performance gap arising from different physics models. At a 0.1 mm aperture, the diffraction spot size significantly exceeds the geometric spot size, allowing the system to adjust its structure to balance aberration and diffraction effects. However, as the aperture increases, the system becomes aberration-limited, reducing the incentive to trade aberration performance for diffraction control. Moreover, compared with the singlet lens, the Cooke triplet and six-asphere designs have higher structural flexibility and hence exhibit more variation in lens configurations. 

\begin{figure}[!t]
     \centering
     \includegraphics[width=0.48\textwidth]{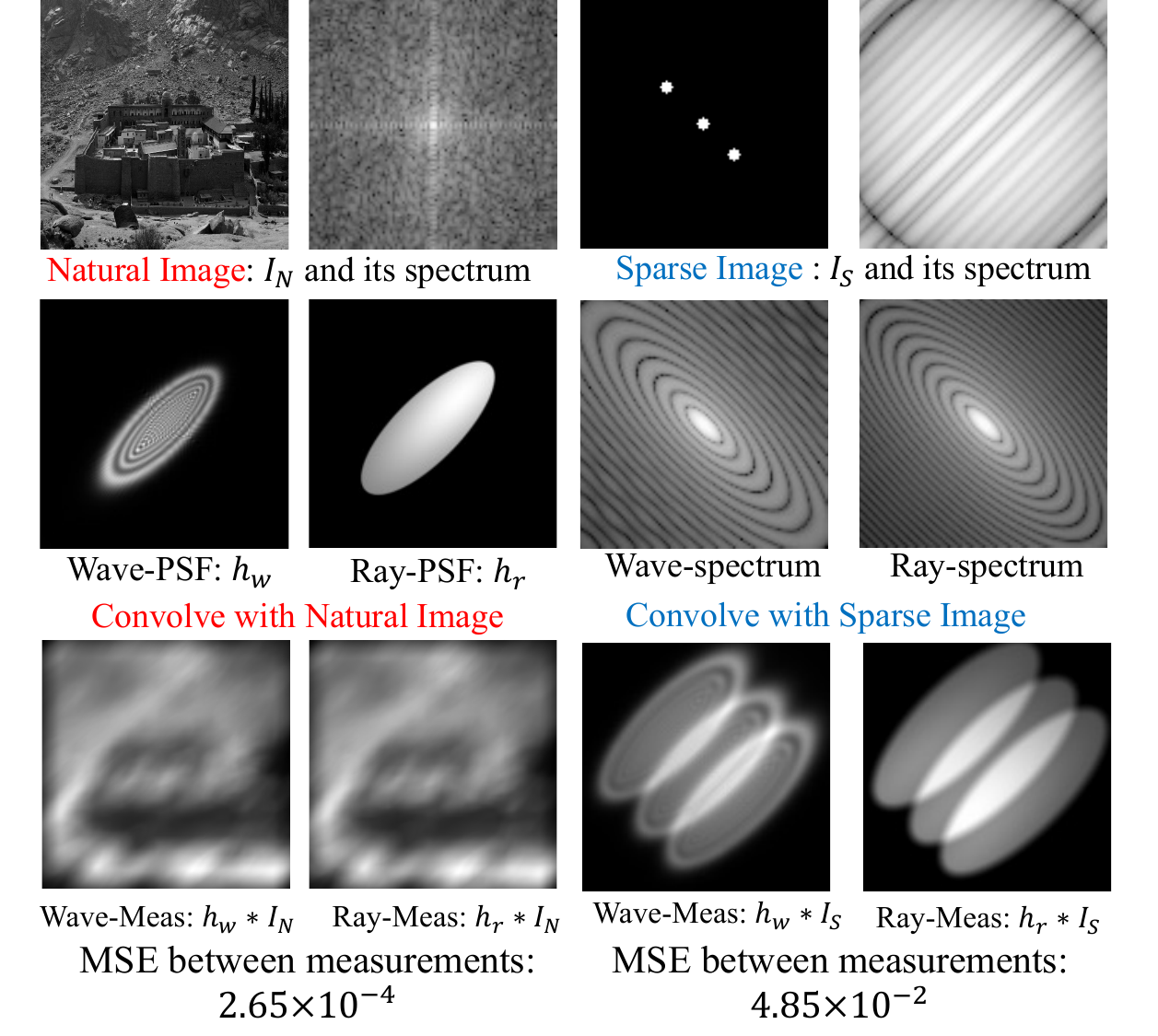}
     \vspace{-6mm}
     \caption{\textbf{Comparing ray and wave measurements in an aberration limited system.} The key spectral difference between ray- and wave-PSFs lies in high frequencies, affecting measurements only when the image has rich high-frequency components. Thus, for natural images, both systems receive similar training data and yield similar configurations. The MSE is measured using normalized measurement intensity.}
     \vspace{-3mm}
     \label{fig:aberration}
\end{figure}

We further investigate the impact of diffraction in the optimization of aberration-limited optics in Fig. \ref{fig:aberration}. The experiments are conducted in a single lens at a 30° off-axis field point with a wavelength of 440 nm. As observed, despite structural differences between wave- ($h_w$) and ray-PSF ($h_r$), their spectra remain similar at low frequencies, where the energy of the natural image $(I_N)$ spectrum is concentrated. Thus, their convolved sub-scenes, $h_w*I_N$ and $h_r*I_N$, exhibit negligible MSE. The MSE is only noticeable between measurements from inputs with rich high-frequency contents, such as $h_w * I_S$ and $h_r * I_S$, which are rare in existing datasets. As a result, with natural imaging datasets and aberration-limited systems, diffraction plays a minor role in end-to-end optimization.

\begin{table}[t]
\caption{The SSIM between sparsely interpolated and reference measurement and time elapsed in interpolation.}
\vspace{-2mm}
\centering
\begin{tabular}{c|c|c|c|c|c}
\hline
\multirow{2}{*}{Lens} & \multirow{2}{*}{FoV} & \multicolumn{4}{c}{Number of PSFs in interpolation}  \\
\cline{3-6}
 & & 9 & 25 & 81 & 289  \\
\hline
\multirow{4}{*}{Singlet} & 5° & 0.987 & 0.990 & 0.995 & 0.999 \\
& 15° & 0.894 & 0.954 & 0.974 & 0.996 \\
& 30° & 0.815 & 0.842 & 0.871 & 0.981 \\
& Time & 7.66 & 12.38 & 36.10 & 96.50 \\
\hline
\multirow{4}{*}{\makecell{Cooke \\ Triplet}} & 5° & 0.995 & 0.995  & 0.996 & 0.999 \\
& 15° & 0.994 & 0.995 & 0.996 & 0.999 \\
& 30° & 0.889 & 0.942 & 0.957 & 0.993 \\
& Time & 7.21 & 9.71 & 25.47 & 62.33 \\
\hline
\multirow{4}{*}{\makecell{Six \\ Aspheric}} & 5° & 0.998 & 0.998 & 0.998 & 0.999 \\
& 15° & 0.996 & 0.996 & 0.997 & 0.997\\
& 30° & 0.998 & 0.998 & 0.998 & 0.999\\
& Time & 8.70 & 13.68 & 37.03 & 104.06 \\
\hline
\end{tabular}
\vspace{-5mm}
\label{table: interp}
\end{table}

\subsection{Interpolation}

To evaluate interpolation accuracy across different FoVs and lens complexities, we use measurements rendered with 969 PSFs, the maximum feasible under hardware limits, as the reference and compare them to interpolated measurements using 9–289 PSFs. As shown in Table \ref{table: interp}, systems with larger FoV require more PSFs to reduce disparity due to stronger aberrations and reduced isoplanaticity. Conversely, systems with more complex lenses exhibit weaker aberrations and thus need fewer PSFs for accurate rendering. Table \ref{table: interp} also lists the computation time for interpolating a single image, showing that denser interpolation significantly increases cost. Thus, selecting the appropriate number of PSFs is critical to balance accuracy and efficiency, depending on FoV and lens complexity.

\subsection{Hardware Validation}

\begin{figure}[!t]
\begin{subfigure}[b]{0.115\textwidth}
\includegraphics[width=19mm]{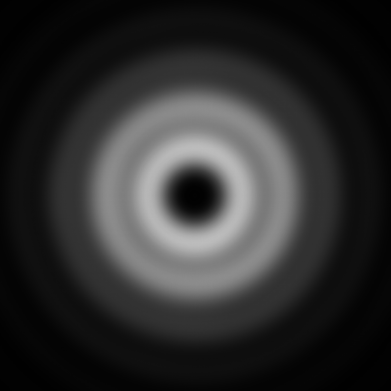}
         \caption{On-Sim.}
         \label{fig:gt_unet}
     \end{subfigure}
     \begin{subfigure}[b]{0.115\textwidth}
    \includegraphics[width=19mm]{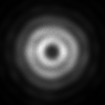}
         \caption{On-Real.}
         \label{fig:gt_unet}
     \end{subfigure}
\begin{subfigure}[b]{0.115\textwidth}
\includegraphics[width=19mm]{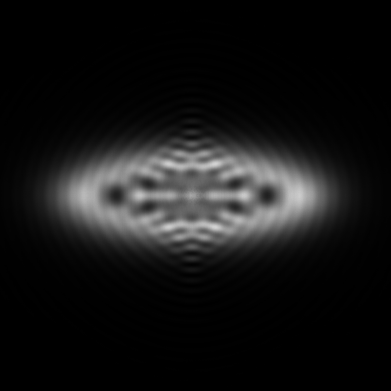}
         \caption{Off-Sim.}
         \label{fig:gt_unet}
     \end{subfigure}
     \begin{subfigure}[b]{0.115\textwidth}
    \includegraphics[width=19mm]{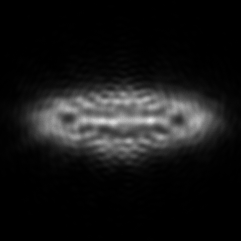}
         \caption{Off-Real.}
         \label{fig:gt_unet}
     \end{subfigure}
     \vspace{-3mm}
     \caption{\textbf{Comparing simulated and real PSFs.} By sending monochromatic parallel beams into a physical lens, we measure real on- and off-axis (15°) PSFs (Real) and compare them with our simulated measurements (Sim.). Our simulator closely matches the real measurements by accurately modeling diffraction and aberration. PSF size: $0.217$ (on) and $0.62$ (off) $\text{mm}^2$.}
     \vspace{-4mm}
     \label{fig:hardware}
\end{figure}

We validate the physical accuracy of our simulator against real-world hardware implementations. In Fig. \ref{fig:hardware}, we send on-axis (0°) and off-axis (15°) parallel monochromatic beams (wavelength: 532 nm) through a plano-convex lens (model 011-1580) onto a sensor (UI-3882LE0M) to generate PSFs, which we then compare with simulated ones. As observed, our simulator accurately models the diffraction patterns and off-axis aberration, yielding similar structures in real and simulated PSFs. The SSIM between real and simulated PSFs are 0.781 (on-axis) and 0.853 (off-axis). These results confirm the reliability of our simulator.

\subsection{Applications}

\begin{figure}[!t]
    \raggedright
    \vspace{0mm}
    \includegraphics[width=0.49\textwidth]{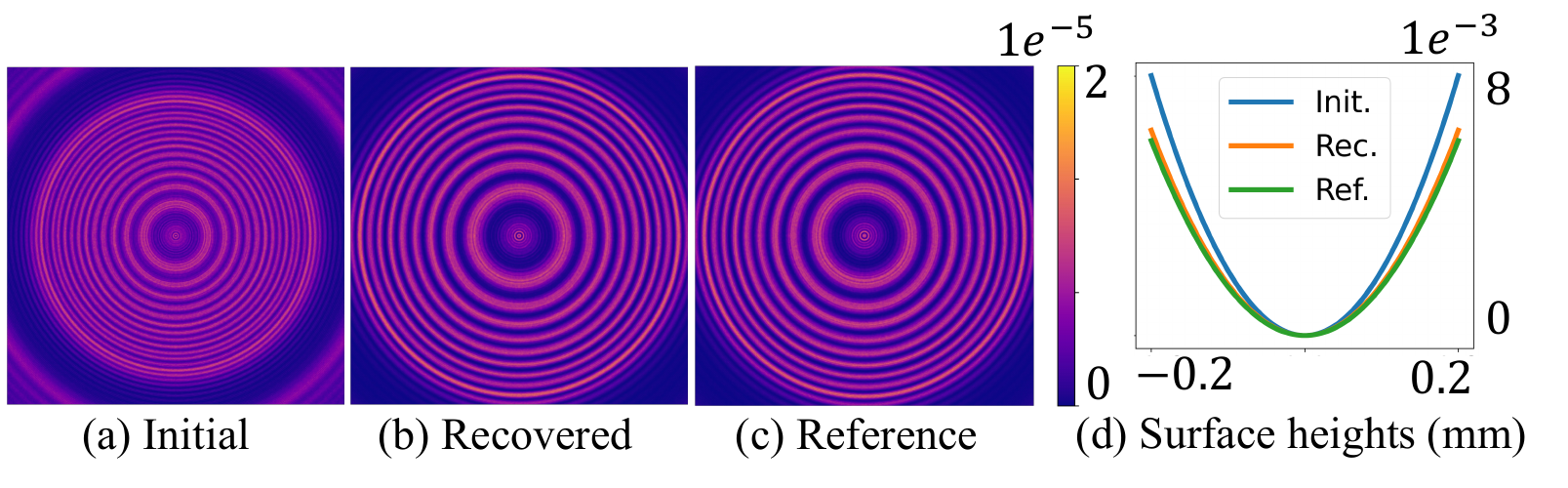}
    \vspace{-3mm}
    \caption{\textbf{Recovering a quadratic surface based on Fizeau interferometer measurements.} Setup: A coherent wavefront is reflected by a quadratic surface, and the resulting interference pattern is detected by the sensor. The interference pattern is determined by the surface geometry. By accurately modeling interference, our differentiable wave optics model results in accurate surface recovery (d). Sensor size: $1.6\text{ mm}^2$.}
    \vspace{-4mm}
    \label{fig:interferometer}
\end{figure}

\noindent \textbf{Fizeau Interferometer:} We simulate Fizeau interferometers \cite{malacara2007optical} as follows: a coherent input wavefront (650 nm) reflects off the test surface, whose profile determines the interference patterns captured by the sensor. Because of the coherency of the light source, all waves should interfere at the sensor plane and hence Eq. \eqref{eq:huygens} becomes a coherent summation of all wavelets from all scene points. The reference measurement is generated with a surface parameterized by reference curvature and quadratic coefficients, as shown in Fig. \ref{fig:interferometer} (c) and (d). We then employ differentiable rendering to recover the surface parameters, initialized with randomly perturbed values (Fig. \ref{fig:interferometer} (d)) with corresponding measurement (Fig. \ref{fig:interferometer} (a)). The optimization is driven by the MSE between the recovered and reference measurements. Because our wave optics model accurately captures phase interference, which reflects surface structures, both the surface (Fig. \ref{fig:interferometer} (d)) and measurement (Fig. \ref{fig:interferometer} (b)) are accurately recovered. This experiment demonstrates the applicability of the proposed model to coherent interference.

\begin{figure}[!t]
    \centering
    \vspace{-2mm}
\includegraphics[width=0.48\textwidth]{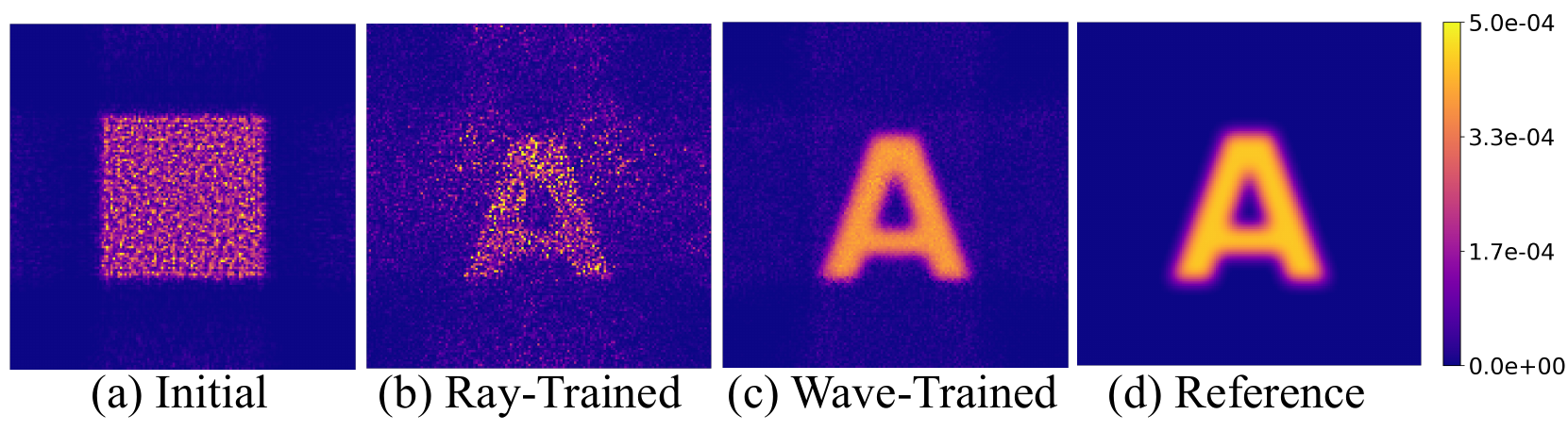}
    \vspace{-7mm}
     \caption{\textbf{Optimizing a freeform optical surface under coherent illumination}. Setup: A monochromatic plane wave is modulated by a freeform optical surface. Due to its coherence, the modulated wavefront interferes with itself in propagation, which can only be accounted for by wave optics. As a result, the wave-trained surface yields accurate recovery, which is not achievable by the ray-trained one. Sensor size: $5.8 \text{ mm}^2$.}
     \label{fig:ff_letter_A}
     \vspace{-6mm}
\end{figure}

\noindent \textbf{Freeform Optics:} We perform differentiable rendering on freeform optics imaging, which is obtained by illuminating the surface with a coherent plane wave (wavelength: 650 nm) and accounting for coherent ray interactions. Specifically, we recover the target measurement in Fig. \ref{fig:ff_letter_A} (d) by surface optimization. The surface is randomly initialized with measurement in Fig. \ref{fig:ff_letter_A} (a), and we conduct ray- and wave-optimization for surface recovery, both are guided by minimizing the MSE between rendered and target measurements. Because of its coherent nature, accounting for wave optics is required for accurate light propagation. Therefore, as shown in Fig. \ref{fig:ff_letter_A} (b) and (c), the recovery is accurate only when wave optics effects are incorporated. These results underscore the versatility and importance of our differentiable wave optics simulator in non-lens optical systems.

\vspace{-1mm}
\subsection{Limitations}

Although our simulator outperforms existing methods, its fidelity and efficiency remain constrained by system scale and aberration strength. For example, large-aperture systems with strong aberrations require very high sampling rates to accurately model wavefronts and propagation \cite{zemax}. Similar issues arise in systems lacking a well-defined focal length or aperture stop, where the wavefront deviates significantly from a nominal sphere. In such cases, using a reference sphere can lead to large residual errors and high sampling demands. Aligning the wavefront by constant OPL may improve sampling efficiency, and exploring such alternative reference geometries is left for future work.
\vspace{-2mm}
\section{Conclusion}

End-to-end optimization leverages the interplay between optics and computational algorithms, but existing frameworks lack the accuracy and efficiency to assess wave optics requirements. We present an efficient, differentiable wave optics simulator that reveals how diffraction impacts joint lens and algorithm design. Experiments show that neglecting diffraction leads to suboptimal configurations and degraded performance under diffraction-limited conditions. These results underscore the need for physics-aware modeling, further validated through differentiable rendering for Fizeau interferometers and freeform optics.
\vspace{-2mm}
\section*{Acknowledgements}
This work was supported in part by the Early Career Faculty Development Award for N. Antipa and T.-M. Li from the Jacobs School of Engineering at UC San Diego, the Ronald L. Graham Chair and the UC San Diego Center for Visual Computing.  We also acknowledge NSF grant 2341952, ONR grant N00014-23-1-2526, and gifts from Adobe, Google, Qualcomm and Rembrand. 

{
    \small
\bibliographystyle{ieeenat_fullname}
    \bibliography{ref}
}



\end{document}